\documentclass[conference]{IEEEtran}
\usepackage{cite}
\usepackage{amsmath,amssymb,amsfonts}
\usepackage{graphicx}
\usepackage{textcomp}
\usepackage{xcolor}
\def\BibTeX{{\rm B\kern-.05em{\sc i\kern-.025em b}\kern-.08em
    T\kern-.1667em\lower.7ex\hbox{E}\kern-.125emX}}
 \usepackage{CJK}
\usepackage{algorithm}
\usepackage{algorithmicx}
 \usepackage{algpseudocode}
\floatname{algorithm}{Algorithm} 
\usepackage{multirow}
\newtheorem{theorem}{Theorem}
\newtheorem{corollary}{Corollary}
\newtheorem{assumption}{Assumption}
\usepackage{booktabs}
\usepackage{microtype}

\begin{document}

\title{Personalized federated learning based on feature fusion
}
\author{
    \IEEEauthorblockN{Wolong Xing$^{a}$, Zhenkui Shi$^{a,b*}$, Hongyan Peng$^{a,b}$,Xiantao Hu$^{a,b}$,Xianxian Li$^{a}$}
    \IEEEauthorblockA{$^a$ Key Lab of Education Blockchain and Intelligent Technology, Ministry of Education,\\
Guangxi Normal University, Guilin, China}
    \IEEEauthorblockA{$^b$ Guangxi Key Lab of Multi-source Information Mining Security,\\ Guangxi Normal University, Guilin, China}
}

\maketitle

\begin{abstract}
 Federated learning enables distributed clients to collaborate on training while storing their data locally to protect client privacy. However, due to the heterogeneity of data, models, and devices, the final global model may need to perform better for tasks on each client. Communication bottlenecks, data heterogeneity, and model heterogeneity have been common challenges in federated learning.
In this work, we considered a label distribution skew problem, a type of data heterogeneity easily overlooked. In the context of classification, we propose a personalized federated learning approach called pFedPM. In our process, we replace traditional gradient uploading with feature uploading, which helps reduce communication costs and allows for heterogeneous client models. These feature representations play a role in preserving privacy to some extent.
 We use a hyperparameter $a$ to mix local and global features, which enables us to control the degree of personalization. We also introduced a relation network as an additional decision layer, which provides a non-linear learnable classifier to predict labels. Experimental results show that, with an appropriate setting of $a$, our scheme outperforms several recent FL methods on MNIST, FEMNIST, and CRIFAR10 datasets and achieves fewer communications.
\end{abstract}


\section{Introduction}
The rapid advances in deep learning have benefited dramatically from large datasets like \cite{001}. However, in the real world, data may be distributed on numerous mobile devices and the Internet of Things(IoT), requiring decentralized training of deep networks.
 Driven by such realistic needs, federated learning \cite{002,003,004} has become an emerging research topic. Federated learning (FL) enables distributed clients to train a generic global model collaboratively on a centralized server while keeping data locally to preserve clients' privacy \cite{1}. 
Due to the advantages of privacy-preserving and training across clients, FL has been widely used and studied in industry and academia. But it still needs to be fully exploited and faces many challenges, such as data heterogeneity, model heterogeneity, and communication bottleneck.

 The most notorious challenge in federated learning is data heterogeneity. Due to differences in regions and behaviors, clients' local data exhibit significant heterogeneity. The current mainstream approach to addressing this challenge is personalized federated learning\cite{4}. Filip Hanzely et al. \cite{6} have proposed a method that generates customized models for each client by mixing local and global models to balance the two. Another commonly used personalized federated learning method is model decoupling, which separates the model into representation layers and decision layers, with the representation layers extracting features from the data and the decision layers making classification predictions based on these features. However, these personalized methods mainly rely on gradient aggregation, which leads to higher communication costs and also limits model heterogeneity.

Mi Luo et al.\cite{005} found that in data-heterogeneous scenarios, the deeper layers of the model exhibit higher heterogeneity in CKA (centered kernel alignment) feature similarity across different clients. The classifier layer has a more significant bias than other layers, and the classifier weight norms tend to favor classes with more training samples. However, the method aims to train a global model to address data heterogeneity. In practice, a single global model only works for some clients. We are inspired by the observation above: under data heterogeneity, significant differences exist in the final layer outputs of the representation layers among different clients. However, these differences benefit personalized federated learning as they reflect locally specific data distributions. We can leverage these feature differences to improve the performance of local models better. Our particular steps are as follows:

First, we consider a heterogeneous scenario with skewed label distribution\cite{003}, which consists of an imbalance in the number of labels and a class-missing method. This paper proposes a new personalized FL scheme(pfedPM) based on feature fusion. In our scheme, we decouple the local client model into two parts: the feature extraction module and the decision module. As shown in Fig\ref{figure:overview}, We designed a two-layer model structure with a shared feature extraction module. We used a conventional neural network as the decision module in the first layer.
In contrast, we employed a relation module from few-shot learning as the decision module in the second layer. Here, the local clients replace the traditional gradient updates by uploading their local data feature representations. These local data features refer to the average output of the feature extraction module for the data with the same label. For each category label, there is a corresponding local feature. This approach allows the client models to be heterogeneous, as we only need to ensure that the dimensions of the feature information are the same. At the same time, the communication cost required for federated learning is positively correlated with the number of categories and the dimensions of the feature information from local clients. Assuming the size of gradients uploaded by the model is 20,000, and the size of feature information is 50, with an average of 10 classes per client. Compared to gradient uploading, the dimensions of our method's upload are 10 * 50, much smaller than the traditional one. This is effective in alleviating communication bottlenecks. For the locally uploaded feature information sent to the server, the server performs average aggregation on features with the same label to obtain the global feature information for that label, which is then sent back to the clients. We use a hyperparameter $a$ to mix local and global features to find the optimal balance. In the local model updates, we introduced a regularization term to ensure that the local feature information stays consistent with the mixed features information. Finally, we conducted comprehensive experiments to evaluate the effectiveness of our approach.
The results show that at appropriate values of $a$, our method outperforms several recent FL methods and achieves less communication on the MNIST, FEMNIST, and CIFAR10 datasets.

In summary, the main contribution is as follows.
\begin{itemize}
    \item We propose a personalized FL based on features fusion which can significantly reduce communication costs and avoid gradient attacks.
    \item We design a hyperparameter  $a$ that works to find the optimal combination of local and global features, allowing us to control the degree of personalization.
    \item We are using relation networks acting as a nonlinear learnable decision module. The experiments demonstrate that relational networks are more effective in finding decision boundaries of different classes in classification tasks.
    \item We conducted comprehensive experiments to evaluate our approach. The results demonstrate that our method outperforms several existing FL methods under appropriate hyperparameter conditions and achieves reduced communication overhead.
     \end{itemize}

\section{related work}

Federated learning \cite{001,003,004} is a rapidly evolving field of research that still has many open questions. This work focuses on solving the non-IID dilemma \cite{006,007} and communication bottlenecks. Related work has followed the following directions.

\subsection{Label Distribution Skew In FL}

In recent years, many studies have focused on analyzing imbalanced data. Real-world data often exhibits imbalanced distributions, which seriously affects the effectiveness of machine learning. Resampling\cite{45} and reweighting\cite{46} are traditional methods for dealing with imbalanced data. Recent work uses reweighting methods to make  
the network pay more attention to minority categories by assigning
variable weights to each class. New perspectives, such as decoupled training\cite{47} and delayed rescaling schemes\cite{48}, have also been proven effective. Most previous work
on imbalanced data has focused on long-tail distributions. However, in the setting of federated learning, data may
be imbalanced in many ways, such as quantity-based 
label imbalance and distribution-based label imbalance. Label distribution bias is always present in practical applications.
For example, pandas are only found in China and zoos, and a person's face may only appear in a few places worldwide. Additionally, label
distribution skew includes long-tail scenarios, but long-tail methods cannot handle the issue of missing classes, which is common in federated learning.
One research strategy to address the statistical heterogeneity of data distribution is to train
multiple global models through clustering and train different global models for
different client groups\cite{2}.
Another approach is to clone the global model, adaptively update high-performing subsets of the global model, and discard poorly performing models to generate specialized models for each client\cite{3}. However, these methods require maintaining multiple global models, which increases communication overhead.
\subsection{Personalized Federated Learning}

Personalized Federated Learning (PFL)\cite{4} is one of the\\ most dominant approaches to solving the
heterogeneity \\problem. 
 Approaches to PFL can be roughly classified according to the strategies used to generate personalized models, such as meta-learning, local and global model mixing, parameter decoupling, clustering, regularization, etc. 
 
  Filip Hanzely et al. \cite{6} have proposed a method that generates personalized models for each client by mixing local and global models to balance the two.
Another common strategy is decoupling the local model into a representation and decision layers. The representation layer extracts features from the data, while the decision layer classifies the extracted features for prediction. 
FedPer \cite{27} learns the body and head during the local training phase and shares only the body with the server. Therefore, the local, personalized model comprises a shared body part and a personalized head. On the other hand, FedRep \cite{28} divides its local updates into two steps. Specifically, each client learns its personalized head by receiving a fixed body from the server, and then it updates the body with the latest personalized head.
FedBABU \cite{29} only learns the body with random initialization and a fixed head during the local update phase and shares only the body with the server. After the training, the global model is fine-tuned to obtain personalized models.
 FedROD \cite{30} designs a dual-head single-body architecture consisting of a general head trained with class-balanced loss and a personalized head trained with experience loss. The body and available head are shared with the server for aggregation, while the personalized head remains private. FedNH \cite{31} combines class prototype consistency and semantics to learn high-quality representations for classification and imposes constraints on the head to prevent prototype collapse due to class imbalance issues. Local model updates are also performed in two steps: first, the body is fixed to update the head, and then the head is fixed to update the body.

 However, the aforementioned personalized federated learning methods are mainly based on gradient uploads, which raises concerns about communication efficiency and gradient attacks \cite{013}.
 
 \subsection{Communication bottlenecks in federal learning}
The issue of communication overhead has been one of the significant challenges in federated learning. In practical scenarios, especially when the required global model size is large, network bandwidth constraints and the number of working nodes can exacerbate communication bottlenecks in federated learning, leading to client devices dropping out and other issues. The most straightforward approach to address the communication overhead issue is to train only low-capacity models\cite{008}. These models occupy less space throughout the federated learning framework, but the trade-off is reduced model accuracy. Using additional computation to reduce the number of communication rounds needed for training has limitations in terms of the number of local time elements performed during local training\cite{009}, and it needs to address the problem of model divergence. Model compression strategies\cite{010} may lead to a loss in long-term accuracy and may not be sufficient to address the reduction of communication time\cite{011} solely.
Moreover, it needs to be determined how these compression strategies handle the statistical challenges associated with highly personalized and heterogeneous data distribution. Nader Bouacida \cite{012} proposed an Adaptive Federated Dropout (AFD) algorithm and combined it with existing compression methods to reduce communication costs, resulting in a 57X reduction in convergence time. However, trimming the global model to reduce communication can lead to a loss of accuracy. Model decoupling methods in personalized federated learning can also facilitate communication to some extent, as they only upload the model parameters of the body part.
\subsection{Model Heterogeneity in Federated Learning}
In practical application scenarios, model heterogeneity is common, as clients often have different hardware and computing capabilities \cite{7}. Knowledge Distillation (KD) based Federated Learning (FL) \cite{8} has been proposed to address this challenge by transferring knowledge from a teacher model to a student model with a different model architecture. However, these methods require an additional common dataset to calibrate the outputs of both the student and teacher models, which increases computational costs. Additionally, the performance of the model is also affected by the distribution differences between the common dataset and the client datasets. Some research has attempted to integrate neural architecture search with Federated Learning \cite{001,003,004}, which enables the discovery of custom model architectures for each group of clients with different hardware capabilities and configurations.

\begin{figure*}

\centering 

\includegraphics[width=1\linewidth]{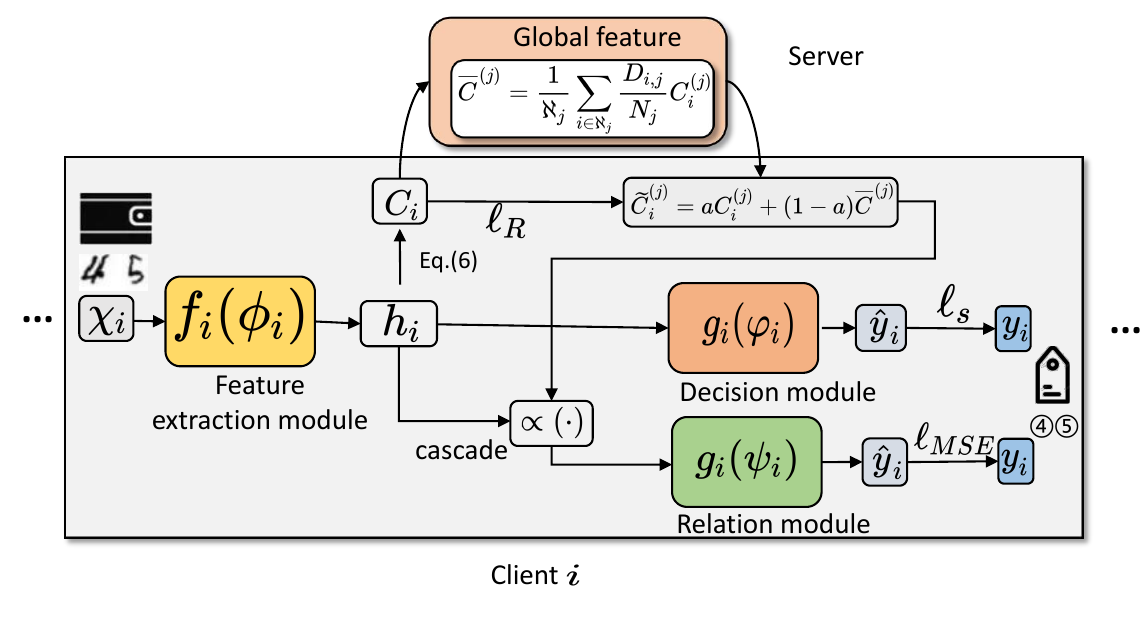} 
\caption{For example, in the heterogeneous environment of pFedPM, the $i$-th client only possesses data with labels 4 and 5. Firstly, the client updates its first-layer model (feature extraction module and decision module) by minimizing the classification loss $\ell_S$ and the distance loss between mixed features and local features $\ell_R$. Next, the feature extraction module is fixed, and the relation module is updated by concatenating mixed features to minimize the loss function $\ell_{MSE}$.} 
\label{figure:overview}
\end{figure*} 

\section{The Personalized FL Scheme based on feature fusion}
\subsection{Previous Federated Learning Setting}
 In conventional Federated Learning (FL), there are $m$ clients communicating with a server to solve the following problem:
 
 \begin{equation}
     \min \limits_{w\in R^d}    f(w) := \dfrac{1}{m}   \sum_{i=1} ^{m}  
                f_{i}(w),
 \end{equation}

\begin{equation}
     f_{i}(w) = E_{{\xi}_{i}} [f_{i}(w ;{\xi}_{i})].
 \end{equation}
  where $f_i(w)$ represents the expected loss on the data distribution of the $i$-th client,  ${\xi}{i}$ is a randomly sampled data point from the data distribution of the $i$-th client, and $f{i}(w;\xi_{i})$ is the loss function corresponding to that sample and $w$. In FL, the data on clients may be non-identical, i.e., the distributions of ${\xi_i}$ and ${\xi_k}$, where $ i \neq k$, can be different. Client data may come from different environments, contexts, and applications.

 However, in the real world, data distributions $P_{i}$ vary among clients, which leads to the global model trained by conventional federated learning only being suitable for some clients. In the context of statistical heterogeneity, $P_{i}$ varies across clients, representing heterogeneous input/output spaces for $x$ and $y$. For example, $P_{i}$ for different clients can be the data distribution over different subsets of classes. In the model heterogeneous setting, $f_{i}$ varies among clients, indicating different model architectures and hyperparameters. The heterogeneity issue can be addressed through personalized federated learning, with the objective function as follows:

 \begin{equation}
     \mathop {\arg \min }\limits_{w_{1},w_{2},w_{3}.....w_{m}} \sum\limits_{i = 1}^m {\frac{{{D_i}}}{N}} (f(w_{i};x),y)  .
 \end{equation}

 where $N$ represents the total data size across all clients, $D_{i}$ represents the data size of the $i$-th client, and $w_{i}$ represents the model of the $i$-th client.

\subsection{Feature fusion-based federated setting}

\label{sec:formatting}

Heterogeneous FL focuses on handling the robustness of heterogeneous input/output spaces, distributions, and model architectures. For instance, the datasets $D_i$ and $D_k$ on two clients, the $i$-th and $k$-th clients may follow different statistical distributions and labels. This is common, for example, in a photo classification app installed on mobile clients. The server needs to recognize many classes $\mathbb{C} = {\mathbb{C}(1), \mathbb{C}(2), …}$, while each client only needs to recognize a few classes that form a subset of $\mathbb{C}$. Although there may be overlaps, the classes can vary across clients. 

According to Figure \ref{figure:overview}, we divide the local model into two parts: the body and the head. The body is primarily used for feature extraction from the data, while the head is responsible for making predictions by identifying different class decision boundaries. We use two different heads (decision module and relation module) that share the same body.

\textbf{Feature extraction module:}  The module is mainly responsible for generating corresponding feature information for input samples with the respective labels. We denote :
\begin{equation}
    h_{i} = f_{i}(\phi _{i},x).
\end{equation}

    where $ h_{i}$ as the feature of the $i$-th client data $x$,$\phi _{i}$ denotes the parameters of the feature extraction module of client $i$.

\textbf{Decision module:}   This module makes a classification decision based on the feature information obtained via the feature extraction module to get the corresponding predicted labels.

\begin{equation}
    f_i(\phi_{i},\varphi_{i}) = g_{i}(\varphi_{i}) \circ f_{i}(\phi_{i}).
\end{equation}

where $\phi_{i}$ represents the parameter of the feature extraction module for client $i$, and $\varphi_{i}$ represents the parameter of the first-layer decision module. We use $w_{i}$ to denote the abbreviation of $(\phi_{i}, \varphi_{i})$.
 
  \textbf{Relation module:} The data input feature extraction module $f_{i}(\phi_i)$ obtains local feature information $h_i$. Then, the local feature $h_i$ is concatenated with the mixed features and entered into the relation module $g_{i}(\psi_{i})$ to calculate the correlation score between them. Finally, the class with the highest correlation is selected as the predicted label.

\begin{figure}

\centering 
\vspace{0cm}
\includegraphics[width=8cm]{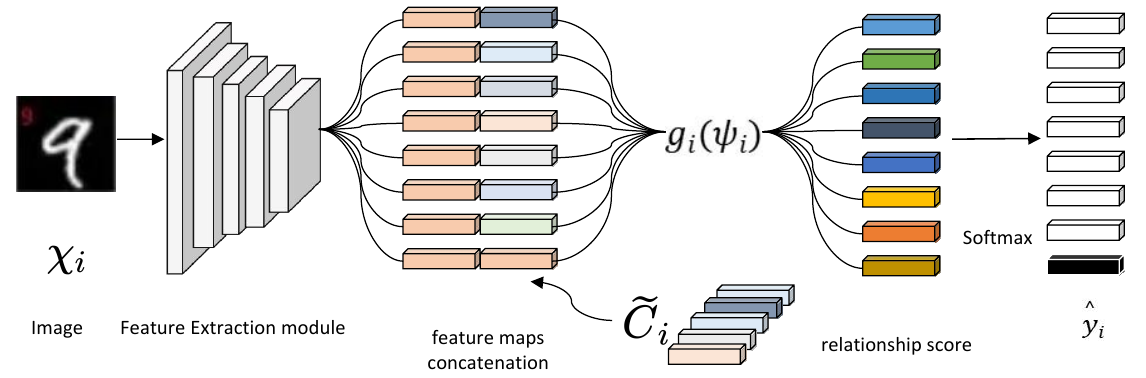} 

\caption{We input an image with a handwritten digit nine into the feature extraction module, obtaining the features of the picture. Since we know this is a ten-classification problem, we first duplicate these ten features. Then, we concatenate them with mixed features corresponding to different labels. These connected features are input into the relation module to obtain correlation scores. Finally, we use a softmax function to get the predicted labels for each feature.} 
\label{fig:wolong2}
\end{figure}

As shown in Figure \ref{fig:wolong2}, assuming the classification task is a 10-class classification problem, the data is transformed into corresponding features $h_i$ using the feature extraction module $f_{i}(\phi_i)$. Then, the features are replicated ten times and concatenated sequentially with the mixed features (for classes missing in the local client, they are connected with the global features). These linked features are then passed to the relation network to obtain similarity scores. Finally, the predicted label for the data is obtained through a softmax function.


\textbf{Local features:} We define a feature $C^{(j)}$ to represent the $j$-th class in $\mathbb{C}$. For the $i$-th client, $C^{(j)}$ is the average of the features obtained by inputting samples with the label $j$ into the feature extraction module.
\begin{equation}
\label{6}
    C_i^{(j)}=\frac{1}{|{D_{i,j}} |} \sum_{(x,y )\in D_{i,j}}f_{i}(\phi_{i};x).
\end{equation}

 where $D_{i,j}$ represents the number of samples of class $j$ in the $i$-th client.

\textbf{Global features:} For a given class $j$, the server receives locally computed features with class label $j$ from a group of clients. These local features with label $j$ are aggregated by taking their average to generate the global feature. $\overline{C}^{(j)}$ for class $j$.

\begin{equation}
 \label{7}
    {\overline C ^{(j)}} = \frac{1}
    {\aleph_j} \sum\limits_{i \in {\aleph _j}} {\frac{|D_{i,j}|}{N_j}} C_{i}^{(j)}.
\end{equation}

where $C_i^{(j)}$ represents the local features of class $j$ from the $i$-th client, $\aleph_j$ represents the set of clients that have class $j$, and $N_j$ represents the total number of
        data samples with class label $j$.

\textbf{Mixed features:}
Data characteristics with the same label may vary across clients, which is common in real-world scenarios. We use a hyperparameter $a$ to train personalized models to mix local and 
    global features. 
Below is the formula for the mixed features of the $i$-th client with class
           label $j$.
\begin{equation}
    \widetilde{C}_i^{(j ) } = a{C}_i^{(j )} + (1 - a){\overline C ^{(j)}}.
\end{equation}
where we use $\widetilde{C}_i^{(j)}$ to represent the mixed features of the $i$-th client with class label $j$, where $a\in[0,1]$ denotes the weight given to the local feature. When $a=0$, it is equivalent to updating the local model using the common global feature. When $a=1$, it is equivalent to using only the local feature. We aim to find an optimal balance between global and local features for each client.



\textbf{Local Model Update:}The client needs to update the local model to generate consistent client features. We also introduced a regularisation to prevent client-side feature drift. Specifically, the loss function is defined as follows:
\begin{equation}
\label{9}
    \ell (D_i,w_i)=\ell _S(f(w_{i};x_i),y_i) +\\
 \lambda \cdot  \ell_R\left ( \widetilde{C}_i^{(j)} ,C_i^{(j)} \right ) .
\end{equation}

where $D_i$ stands for data from the $i$-th client, $\lambda$ is an important parameter for regularization.

\textbf{Optimization Objective}

 We now introduce our new formulation of the FL model guided by training in the decision module.
\begin{equation}
\begin{aligned}
\mathop {\arg \min }\limits_{\phi, \varphi }  \sum\limits_{i = 1}^m \frac{{{D_i}}}{N}\ell _S(f(w_{i};x_i),y_i) +\\
 \lambda\cdot \sum\limits_{j = 1}^{|\mathbb{C} |} \sum\limits_{i = 1}^m \frac{|D_{i,j}|}{N_j} \ell_R\left ( \widetilde{C}_i^{(j)} ,C_i^{(j)} \right ).
\end{aligned}
\end{equation}
 
where loss $\ell _S(f(w{i};x_i), y_i)$ represents the objective loss for the $i$-th client, and we use the standard cross-entropy loss as the objective loss function. $\mathbb{C}$ represents the number of classes for the labels, and $\ell_R$ is the regularization term used to measure distance, with its expression as follows.
\begin{equation}
    \ell _R\left ( \widetilde{C}_i^{(j)} ,C_i^{(j)} \right )=||{\widetilde {{C_{\rm{i}}}}^{(j)}},C_{\rm{i}}^{(j)}||_2 .
\end{equation}

$\ell _R\left ( \widetilde{C}_i^{(j)}, C_i^{(j)} \right )$ corresponds to the regularization term that ensures the model's generalization performance. It measures the distance between local features and mixed features. $N$ is the total number of instances across all clients, $D_{i}$ represents the data size of the $i$-th client, and $N_{j}$ is the number of instances belonging to class $j$ across all clients. $D_{i,j}$ is the number of instances belonging to class $j$ in the $i$-th client.

 The optimization problem can be solved by iterating the following two steps of alternating minimization: (1) minimizing w.r.t. each $w_{i}$ with ${\widetilde {{C_{\rm{i}}}}^{(j)}}$ fixed; and (2) minimizing w.r.t. ${\widetilde {{C_{\rm{i}}}}^{(j)}}$ with all $w_{i}$ fixed. Further details concerning these two steps can be seen in Algorithm \ref{algorithmic1}.
 \begin{CJK*}{UTF8}{gkai}
\begin{algorithm}

    \caption{pfedPM} 
        \begin{algorithmic}[1] 
            \Require $D_i$,$w_i$,$\psi _i$,$i=1,...m$
            \Ensure $\{C_i\}$  ,$i=1,...m$
            \State Initialize global features set $\{\overline C ^{(j)}\}$ for all classes.
            \For{each round $T=1,2,....$} 
                \For{each client $i$ in parallel }     
                    \State $\{C_i\}\longleftarrow$ LocalUpdate $        ( i,\overline C ^{(j)} )$ 
                    
                \EndFor
                \State Update global features by Eq.\ref{7}
                \State Update local features set $C_i$ with prototypes in $\{\overline C ^{(j)}\}$
            \EndFor
            \For{each local epoch }
                \For{batch  $(x_i,y_i)\in D_i$}
                \State Compute local features by Eq.\ref{6}
                 \State Compute loss by Eq.\ref{9} using local features.
                  \State Update local model according to the loss. 
                  \State Update Relationship module according to the  loss\\ by     $    \psi_i\gets \arg \min\sum\limits_{s = 1}^{D_i }\sum\limits_{j = 1}^{|\mathbb{C} | }\ell_{MSE}  (R^{(i)}_{s,j}-1(y_s==y_j))^2$
                \EndFor{}
            
            \EndFor{}
            \State \Return $C_i$ 
        \end{algorithmic}
        \label{algorithmic1}
        
    \end{algorithm}
\end{CJK*}

Next, we introduce the new formula for training the model under the relation module.

Unlike everyday classification tasks that use cross-entropy loss functions, we use mean squared error to supervise the similarity scores. The optimization objective function is as follows:

\begin{equation}
    \psi_i\gets \arg \min\sum\limits_{s = 1}^{D_i }\sum\limits_{j = 1}^{|\mathbb{C} | }\ell_{MSE}  (R^{(i)}_{s,j}-1(y_s==y_j))^2,
\end{equation}

and
\begin{equation}
   R^{(i)}_{s,j}=
g_{\psi_{i}}(\propto  (f_{\phi_{i}}(x_s),\widetilde{C}_i^{(j)} )  ),j\in \mathbb{C}  . 
\end{equation}

 \addtolength{\topmargin}{-.13in}
where $f_{\phi}(\cdot)$ represents the feature information after passing through the feature extraction module, $\propto$ denotes the concatenation operation, and for the mixed features of the $i$-th client with the missing class, we use the global feature as a replacement. $R^{(i)}_{s,j}$ represents the correlation score between the $s$-th data of the $i$-th client and the mixed features $\widetilde{C}_i^{(j)}$. $y_s$ is the true label of the $s$-th data,  $y_j$ is the label of the mixed features $\widetilde{C}i^{(j)}$ and $g{\psi}(\cdot)$ is the correlation computation function.
\addtolength{\topmargin}{-.007in}
\addtolength{\topmargin}{-.009in}
\addtolength{\topmargin}{-.09in}
\section{Convergence analysis}
We use the first-level model (decision module) as our objective loss function.
\begin{assumption}
(Lipschitz Smooth). It is assumed that each local objective function is $L_1$-Lipschitz Smooth, which also implies that the gradient of the local objective function is $L_1$-Lipschitz continuous.
\label{assumption1}
\end{assumption} 

\begin{equation}
    \begin{aligned}
         \left \|  \bigtriangledown \ell_ {t_{2}} -\bigtriangledown\ell_{t_{2}}  \right \|_{2}\le
 L_1 \left \| w_{i,t_{1}}-w_{i,t_{2}} \right \|_{2},\\ \forall t_1,t_2>0,i\in \{1,2,....m\}.
    \end{aligned}
\end{equation}

This also implies the following quadratic bound,
\begin{equation}
    \begin{aligned}
        \ell_ {t_{1 }} -\ell_{t_{2}}  \le  \left \langle \bigtriangledown \ell_ {t_{2}},(w_{i,t_{1}}-w_{i,t_{2}}) \right \rangle +\\ \frac{L_1}{2}  \left \| w_{i,t_{1}}-w_{i,t_{2}} \right \|_{2}^2 \forall t_1,t_2>0,i\in \{1,2,....m\}.\\
    \end{aligned}
\end{equation}

\begin{assumption}
(Unbiased Gradient and Bounded Variance)
The stochastic gradient $g_{i,t}=\ell(w_{i,t})$ is an unbiased estimator of the local gradient for each client. Assuming that its expectation satisfies the following equation,
\label{assumption2}
\end{assumption} 
         
\begin{equation}
    \begin{aligned}
        E_{{\xi}_i}\sim _{D_i}\left [ g_{i,t} \right ] =\bigtriangledown \ell(w_{i,t})=\bigtriangledown \ell_t,\\\forall i\in \{1,2,...m\},
    \end{aligned}
\end{equation}

and its variance is bounded by  $ \sigma ^2$:
\begin{equation}
    E \left [ \left \| g_{i,t}-\bigtriangledown \ell_{(w_{i,t})} \right \| _2^2 \right ]  \le \sigma ^2.
\end{equation}

\begin{assumption}\label{assumption3}
(Bounded Expectation of Euclidean norm of Stochastic Gradients). The expectation of the random gradient is bounded by $G$:
\end{assumption} 
\begin{equation}
    E \left [ \left \| g_{i,t} \right \| _2 \right ]  \le G,\forall i\in \{1,2,...m\}.
\end{equation}

\begin{assumption}\label{assumption4} The functions of each feature extraction module, commonly known as embedding functions, are $L_2$-Lipschitz continuous.
\end{assumption}

\begin{equation}
    \begin{aligned}
        \left \| f_i(\phi _{i,t_1})- f_i(\phi _{i,t_2})  \right \|\le 
L_2\left \|\phi _{i,t_1}-\phi _{i,t_2}  \right \| _2,\\\forall t_1,t_2>0, i\in \{1,2,...m\}.
    \end{aligned}
\end{equation}

We can obtain theoretical results for non-convex problems if the above assumption holds. In Theorem \ref{theorem1}, we provide the expected decrease in each round. We use $  e \in \{ 1/2, 1, 2, . . . , E \}$ to represent local iterations and $t$ to represent global communication rounds. Here, $tE$ represents the time step before global features aggregation, and $tE + 1/2$ represents the time step between global features aggregation and the first iteration of this round.


\begin{theorem}\label{theorem1}
(One-round deviation) Let Assumption \ref{assumption1} to \ref{assumption4} hold. For an arbitrary client, after every communication round, we have,
\end{theorem} 

\begin{equation}
    \begin{aligned}
        \left [ \left \| \ell _{(t+1)E+1/2} \right \|  \right ] \le \ell _{tE+1/2}-\left ( \eta -\frac{L_{1} \eta^2}{2}  \right ) \\\sum_{e=1/2}^{E-1} \left \| \bigtriangledown \ell _{tE+e} \right \| _2^2+\frac{L_{1}E \eta^2}{2}\sigma ^2+\lambda L_2 \eta EG.
    \end{aligned}
\end{equation}

Theorem \ref{theorem1} indicates the deviation bound of the local objective function for an arbitrary client after each communication round. Convergence can be guaranteed when there is a
certain expected one-round decrease, which can be achieved
by choosing appropriate $\eta$ and $\lambda$ .

\begin{corollary}\label{corollary1}
(Non-convex pFedPM convergence). The
loss function $\ell$ of an arbitrary client monotonously decreases in every communication round when
\end{corollary} 

\begin{equation}
    \begin{aligned}
        \eta_{\acute{e}}  < \frac{2( {\textstyle \sum_{e=1/2}^{\acute{e}}} \left \|\bigtriangledown \ell _{tE+e}\right \| _2^2-\lambda L_2EG ) }{L_2  EG} ,
    \end{aligned}
\end{equation}

     where  $\acute{e} = \{ 1/2, 1, 2, . . . , E \}$.and \begin{equation}
    \lambda_t < \frac{\left \|\bigtriangledown \ell _{tE+1/2}\right \| _2^2  }{L_2  EG} .
\end{equation}

Thus, the loss function converges .

Corollary \ref{corollary1} guarantees that the expected bias of the loss function is negative, ensuring the convergence of the loss function. We can further ensure the convergence of the algorithm by choosing appropriate learning rates $\eta$ and importance weights $\lambda$.
\section{Experiments}
\textbf{Privacy Preserving}
\addtolength{\topmargin}{-.001in}
 
 Our approach involves exchanging feature information of labels between the server and client machines instead of model parameters, which is beneficial for privacy protection. First, the low-dimensional representation of labels from the same category is averaged to generate feature information as a one-dimensional vector, and this process is irreversible. Secondly, even with reconstructing uploaded gradients for the original data through gradient attacks, it's impossible to reconstruct the actual client data from the feature information. This effectively safeguards the privacy of the clients.

\begin{table*}[t]

    \centering
     \resizebox{\linewidth}{!}{
    \begin{tabular}{cccccccc}
     \toprule
        DataSet & Method & Stdev &  Test Average Acc n=3 & Test Average Acc n=4 & Test Average Acc n=5 &  Rounds & Params ($\times$1000)\\
 \midrule
        MNIST & Local & 2 &  94.05$\pm$2.93 &  93.35$\pm$3.26 &  92.92$\pm$3.17 &  0 & 0 \\ 
        MNIST & FeSEM & 2 &  95.26$\pm$3.48 & 97.06$\pm$2.72 & 96.31$\pm$2.41 & 150 & 430 \\ 
        MNIST & FedProx & 2 &  96.26$\pm$2.89 &  96.40$\pm$3.33 &  95.65$\pm$3.38 & 110 & 430 \\ 
        MNIST & FedPer & 2 &  95.57$\pm$2.96 &  96.44$\pm$2.62 &  95.55$\pm$3.13 & 100 & 106 \\ 
        MNIST & FedAvg & 2 &  95.04$\pm$6.48 &  94.32$\pm$4.89 &  93.22$\pm$4.39 & 150 & 430 \\ 
        MNIST & FedRep & 2 &  94.96$\pm$2.78 &  95.18$\pm$3.80 &  94.94$\pm$2.81 & 100 & 110 \\ 
        MNIST & L2GD & 2 &  97.42$\pm$2.78 &  96.38$\pm$1.56 &  95.86$\pm$2.42 & 100 & 110 \\ 
        MNIST & pFedPM & 2 &\textbf{98.50$\pm$0.42} & 97.52$\pm$0.51 & 96.84$\pm$0.24 & 100 & 4 \\ 
        MNIST & pFedPM-mh & 2 & {98.24$\pm$0.43} &\textbf{ 97.70 $\pm$0.52}& \textbf{96.89 $\pm$0.26}& 100 & 4 \\ 
        \midrule 
        FEMNIST & Local & 1 &  92.50$\pm$10.42 &  91.16$\pm$5.64 &  87.91$\pm$8.44 & 0 & 0 \\
        FEMNIST & FeSEM & 1 &  93.39$\pm$6.75 & 91.06$\pm$6.43 &  89.61$\pm$7.89 & 200 & 16000 \\ 
        FEMNIST & FedProx & 1 & 94.53$\pm$5.33 &  90.71$\pm$6.24 &  91.33$\pm$7.32 & 300 & 16000 \\ 
        FEMNIST & FedPer & 1 &  93.47$\pm$5.44 &  90.22$\pm$7.63 &  87.73$\pm$9.64 & 250 & 102 \\ 
        FEMNIST & FedAvg & 1 &  94.50$\pm$5.29 &  91.39$\pm$5.23 &  90.95$\pm$7.22 & 300 & 16000 \\ 
        FEMNIST & FedRep & 1 & 93.36$\pm$5.34 &  91.41$\pm$5.89 &  89.98$\pm$6.88 & 200 & 102 \\
        FEMNIST & L2GD & 1 &  94.28$\pm$2.24 &  93.78$\pm$2.46 &  91.42$\pm$1.88 & 100 & 110 \\ 
        FEMNIST & pFedPM & 1 & \textbf{96.93$\pm$2.24} &\textbf{95.81$\pm$1.64} & \textbf{93.35$\pm$2.54} & 120& 4 \\ 
         FEMNIST & pFedPM-mh & 1 & 96.85$\pm$1.96 & 94.91$\pm$1.42 & 93.19$\pm$2.18 &120 & 4\\ 
  \midrule 
        CIFAR10 & Local & 1 &  42.79$\pm$8.64 &  42.12$\pm$6.38 &  40.74$\pm$6.52 & 0 & 0 \\
         CIFAR10  & FeSEM & 1 &  56.43$\pm$6.54 & 55.41$\pm$7.64 &  54.32$\pm$5.36 & 300 & 175000 \\ 
         CIFAR10  & FedProx & 1 & 61.58$\pm$4.12 &  61.17$\pm$7.54 &  60.34$\pm$6.58 & 300 & 175000 \\ 
         CIFAR10  & FedPer & 1 &  59.66$\pm$4.32 &  58.44$\pm$6.98 &  57.32$\pm$7.14 & 250 & 156000 \\ 
         CIFAR10  & FedAvg & 1 &  62.56$\pm$4.79 &  61.44$\pm$4.21 &  59.68$\pm$4.35 & 300 & 175000 \\ 
         CIFAR10  & FedRep & 1 & 56.76$\pm$4.28 &  55.43$\pm$6.92 &  55.17$\pm$7.42 & 250 & 175000 \\
         CIFAR10  & L2GD & 1 & 63.42$\pm$5.58 &  62.71$\pm$5.43 &  62.16$\pm$7.53 & 300 & 235000 \\
         CIFAR10  & pFedPM & 1 & 65.42$\pm$4.25 &\textbf{64.76$\pm$4.21} & \textbf{63.86$\pm$3.58} & 250& 82 \\ 
         CIFAR10 & pFedPM-mh & 1 & \textbf{65.76$\pm$3.65} & 64.34$\pm$5.23 & 63.25$\pm$4.18 &250 & 82\\ 
        
    \bottomrule
    \end{tabular}
    }
    
    \caption{\centering The comparison of the FL method on three benchmark datasets is presented in the table, with the best results highlighted in bold. In contrast to the baseline, we have experimentally demonstrated that accuracy can be improved by varying the hyperparameter $a$ for our method.}
    \label{11}
\end{table*}

\textbf{Local models and datasets}

We used three popular benchmark datasets: MNIST \cite{41}, FEMNIST \cite{42}, and CIFAR10 \cite{014}. For the MNIST and FEMNIST datasets, our base model consists of two convolutional layers and two fully connected layers. Specifically, the two convolutional layers serve as our shared feature extraction module, while the two fully connected layers act as the first prediction module. Additionally, we employed one convolutional layer and two fully connected layers as the relationship module. As for the CIFAR10 dataset, we adopted the ResNet18 \cite{015} model.

\textbf{Heterogeneous scene setting}

Each client learns a supervised learning task. We use "n" to control the number of classes and "k" to control the number of training instances per class. We randomly vary the "n" and "k" values among clients to simulate heterogeneous scenarios with skewed label distributions. We define the average values of "n" and "k" and then add random noise to each user's "n" and "k". The purpose of introducing variance in "n" is to control the heterogeneity in the class space, while the purpose of introducing conflict in "k" is to control the data size imbalance.
We set up 20 clients on the MNIST, FEMNIST, and CRIFAR10 datasets, and each client has an average data size of 100 for each class. The batch size is 10, and the number of local update rounds is 1. The local clients are trained with the SGD optimizer, with a learning rate 0.01 and momentum of 0.5. Our initial hyperparameter set is directly from the default hyperparameter set \cite{43}. We assume that all clients perform learning tasks with heterogeneous statistical distributions. To simulate different degrees of heterogeneity, we fix the standard deviation to be 1 or 2.
\addtolength{\topmargin}{-.12in}
\addtolength{\topmargin}{0.22in}
\textbf{Model heterogeneity simulations}

We consider slight differences in the model architecture among clients for the model heterogeneity setting. In MNIST and FEMNIST, the output channel numbers of the convolutional layers are set to 18, 20, or 22, while in CIFAR10, the stride of the convolutional layers is set differently across different clients.

\textbf{Baselines of FL}

We investigated the performance of pFedPM under statistical and model heterogeneity settings (pFedPM-mh), as well as several baselines, including local models trained on each client, FedAvg \cite{1}, FedProx \cite{24}, FeSEM \cite{23}, FedPer \cite{25}, FedRep \cite{26} and L2GD\cite{6}.

  \textbf{$\lambda$   Settings}
  
 For the vital hyperparameter  $\lambda$, we adjust the optimal  $\lambda$ from a limited set of candidates by grid search. the optimal  $\lambda$ values for MNIST, FEMNIST and CIFAR10 are 1, 1 and 0.1.

\begin{figure}

\centering 

\includegraphics[width=8cm]{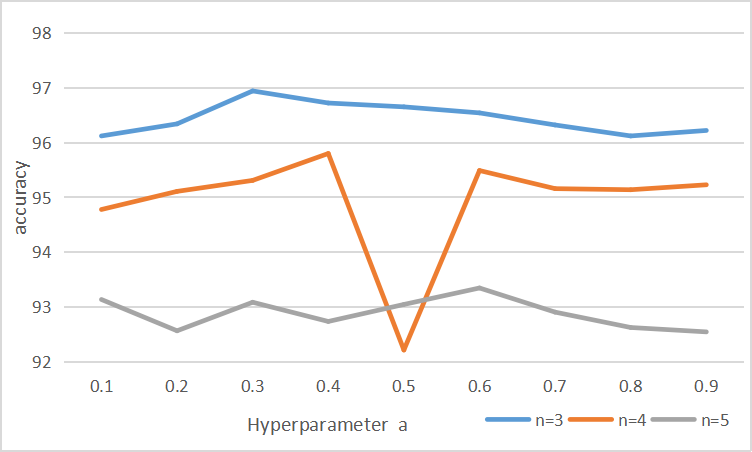} 

\caption{\centering On the FEMNIST dataset, the accuracy corresponding to different hyperparameters $a$} 
\label{99}
\end{figure} 

  \label{sec:intro}

\label{sec:pagestyle}

\begin{table*}[!ht]

    \centering
     \resizebox{\linewidth}{!}{
    \begin{tabular}{cccccccc}
     \toprule
        DataSet & Method & Stdev &  Test Average Acc n=3 & Test Average Acc n=4 & Test Average Acc n=5 &  Rounds & Params ($\times$1000)\\
 \midrule
       
        MNIST & pFedPM & 2 &98.50$\pm$0.42& 97.52$\pm$0.51 & 96.84$\pm$0.24 & 100 & 4 \\ 
        MNIST & pFedPM-mh & 2 & {98.24$\pm$0.43} &{ 97.70 $\pm$0.52}& 96.89 $\pm$0.26& 100 & 4 \\ 
         MNIST & R-pFedPM & 2 &{98.11$\pm$0.52} & 97.73$\pm$0.44 & \textbf{97.01$\pm$0.18} & 100 & 4 \\ 
        MNIST & R-pFedPM-mh & 2 & \textbf{{98.54$\pm$0.25}} &\textbf{{ 97.93 $\pm$0.16}}& {96.68 $\pm$0.46}& 100 & 4 \\ 
  \midrule 
       
        FEMNIST & pFedPM & 1 & 96.93$\pm$2.24 &{95.81$\pm$1.64} & 93.35$\pm$2.54 & 120& 4 \\ 
         FEMNIST & pFedPM-mh & 1 & 96.85$\pm$1.96 & 94.91$\pm$1.42 & 93.19$\pm$2.18 &120 & 4\\ 
          FEMNIST & R-pFedPM & 1 &\textbf{ 97.28$\pm$2.42} &\textbf{{96.31$\pm$1.88}} & 93.12$\pm$2.48 & 120& 4 \\ 
         FEMNIST & R-pFedPM-mh & 1 & 97.10$\pm$2.04 & 95.22$\pm$1.68 & \textbf{93.72$\pm$2.64 }&120 & 4\\ 
         \midrule
         CIFAR10  & pFedPM & 1 & 65.42$\pm$4.25 &64.76$\pm$4.21 & 63.86$\pm$3.58& 120&82 \\ 
         CIFAR10 & pFedPM-mh & 1 & 65.76$\pm$3.65 & 64.34$\pm$5.23 & 63.25$\pm$4.18 &120 & 82\\ 
         CIFAR10  & R-pFedPM-mh& 1 & 66.31$\pm$3.42 &\textbf{65.14$\pm$5.34} & \textbf{64.13$\pm$4.12} & 120& 82 \\ 
         CIFAR10 & R-pFedPM-mh & 1 & \textbf{66.17$\pm$4.35} & 64.74$\pm$6.33 & 63.78$\pm$4.56 &120 & 82\\ 
         
    \bottomrule
    \end{tabular}
    }

    \caption{\centering The relational network is used as a module for predicting labels, and we can find a slight benefit of the relational module( R-pFedPM ) in deciding the classification boundaries of the different categories.}
    \label{22}
\end{table*}
 
 \textbf{$a$ Settings}
 
 We controlled the level of personalization by performing a grid search with different values of parameter $a$ on the MNIST and FEMNIST datasets. As shown in Figure \ref{99}, for the homogeneous models on the FEMNIST dataset, we found that our optimal parameter values were set to 0.3, 0.4, and 0.6 as the heterogeneity increased. We also set up homogeneous and heterogeneous scenarios with different hyperparameters for the three datasets. The results showed that our hyperparameter values increased with the degree of class imbalance.
 
 \textbf{Performance in non-IID Union environments at the setting of the hyperparameter $a$.}
 
As shown in Table \ref{11}, we compared pFedPM with other baseline methods, particularly L2GD (Mixing of Local and Global Models). By setting the appropriate hyperparameter $a$, we can observe that our feature fusion approach outperforms other baseline methods in most cases. The experimental results indicate that the feature fusion method outperforms the model fusion method in terms of accuracy compared to L2GD, requiring lower communication costs. This suggests combining local and global features through feature fusion can improve the model's performance and reduce the data transmission overhead between clients and the server in federated learning. Therefore, feature fusion is a more practical approach to address the heterogeneity problem in federated learning.


 \textbf{Communication efficiency}
 
As shown in Table \ref{11}, we reported the number of communication rounds required for convergence and the parameters communicated in each round in Table 1. It can be observed that our method requires less communication compared to other baselines. The comparison with FedAVG reduced the amount of communication overhead by more than 100 times, indicating a significant improvement in communication efficiency achieved by our approach.

 \textbf{The performance of using the relation module for label prediction in a non-IID environment.}
 
 Clients retain the mixed features from the last round of training during the testing phase and perform classification predictions using the relation module. As seen in Table \ref{22}, in this scenario, the relation module for label prediction outperforms directly using a fully connected neural network for prediction, achieving better results.

\section{Conclusion}
\label{sec:majhead}

This paper proposes a novel feature fusion federated learning (FL) approach to address challenging FL scenarios with heterogeneous input/output spaces, data distributions, and model architectures. The method achieves personalized federated learning by exchanging and blending local and global features. Additionally, we employ a relationship module to identify decision boundaries between different classes, resulting in slight improvements in the experimental results. Based on the empirical findings, our approach outperforms several recent FL methods on the MNIST, FEMNIST, and CIFAR10 datasets.
{\small
\bibliographystyle{IEEEtran}
\bibliography{egbib}
}

\end{document}